%% file: main.tex
\documentclass{article}
\PassOptionsToPackage{dvipsnames,table}{xcolor}
\usepackage{arxiv}
\usepackage{graphicx}
\usepackage[numbers]{natbib}
\usepackage{doi}
\usepackage[utf8]{inputenc} 
\usepackage[T1]{fontenc}    
\usepackage{url}            
\usepackage{booktabs}       
\usepackage{amsfonts}       
\usepackage{nicefrac}       
\usepackage{microtype}      
\usepackage{amsmath}
\usepackage{amssymb}
\usepackage{bbm}
\usepackage{wasysym}
\usepackage{pifont}
\usepackage{threeparttable}
\usepackage{subcaption}
\usepackage{array}
\usepackage{makecell}
\usepackage{algpseudocode} 
\usepackage{multirow}
\usepackage{listings}
\usepackage{enumitem}
\usepackage{hyperref}
\usepackage{todonotes}
\usepackage{comment}
\usepackage[capitalise, noabbrev]{cleveref}
\usepackage{lscape}
\usepackage{tabularray}
\usepackage{CJK}

\usepackage{colortbl}
\usepackage{wrapfig}
\usepackage{adjustbox}
\usepackage{verbatim}
\usepackage{minitoc}
\usepackage{algorithm}
\usepackage[most]{tcolorbox}

\newtcolorbox{abstractbox}{
    enhanced,
    colback=gray!5,
    colframe=black,
    boxrule=0.6pt,
    arc=2mm,
    left=8pt,
    right=8pt,
    top=6pt,
    bottom=8pt,
    before skip=12pt,
    after skip=12pt
}

\definecolor{bestgreen}{RGB}{208,124,176}
\definecolor{secondgreen}{RGB}{235,170,210}
\definecolor{thirdgreen}{RGB}{255,225,245}
\definecolor{bestpurple}{RGB}{120,70,255}

\title{ReDT: \underline{Re}solution-\underline{D}ecoupled
Visual Geometry \underline{T}ransformer}

\author{%
  Abteen Arab$^{1,2,*}$ \quad Guile Wu$^{1}$ \quad Dongfeng Bai$^1$\\
  $^{1}$Huawei Noah's Ark Lab \quad $^{2}$University of British Columbia\\
  \texttt{aarab02@student.ubc.ca, guile.wu@outlook.com, baidongfeng@huawei.com} \\
}

\renewcommand{\headeright}{Technical Report}

\begin{document}

\maketitle

\begin{figure}[h]
    \centering
    \includegraphics[width=0.99\textwidth]{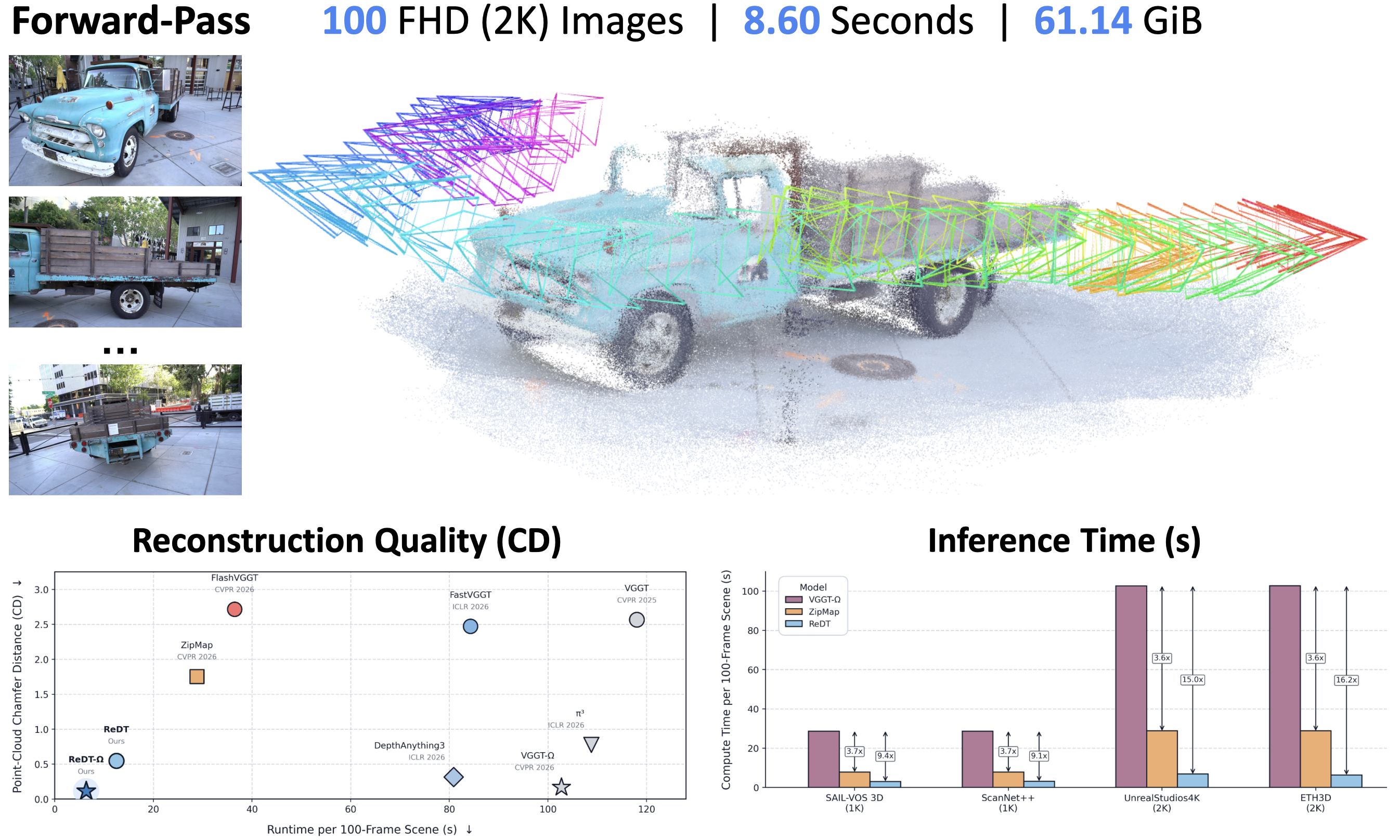}\par
    \captionof{figure}{ReDT achieves state-of-the-art point-cloud reconstruction on ETH3D using 100 input views at 2K resolution (left), while enabling fast inference across HD benchmarks (right).}
    \label{fig:main}
\end{figure}

\begingroup
\renewcommand\thefootnote{*}
\footnotetext{Abteen Arab contributed to this work during an internship at Huawei Canada.}
\endgroup

\thispagestyle{fancy}

\input{Sections/Abstract}
\input{Sections/Introduction}
\input{Sections/Related_work}
\input{Sections/Method}
\input{Sections/Experiments}

\input{Sections/Conclusion}

\newpage

{
\bibliographystyle{plainnat}
\bibliography{ref}
}

\newpage
\addcontentsline{toc}{section}{Appendix}

\input{Sections/Appendix}

\end{document}

%% file: Sections/Abstract.tex
\begin{abstract}
We introduce \emph{ReDT}, a feed-forward visual geometry transformer designed to scale multi-view reconstruction to high-definition (HD) inputs.
Existing visual geometry transformers are typically trained at fixed, standard definition (SD) resolutions and exhibit degraded reconstruction quality and poor computational scaling to high input resolutions.
While recent efficient architectures reduce the complexity of inter-frame attention, expensive intra-frame and encoder attention remain largely unaddressed, making HD reconstruction impractical with existing architectures.
\emph{ReDT} addresses both limitations with a resolution-aware architecture designed to scale to high-resolution inputs.
The key idea of \emph{ReDT} is to first map each view to a fixed-size set of basis tokens, compute correspondences in this compact constant-size basis space, and then project the outputs back to the original resolution for dense prediction.
By decoupling the internal representation from input resolution during attention, \emph{ReDT} substantially reduces the computational cost of high-resolution 3D reconstruction and improves model performance at all resolutions.
Extensive experiments on multiple datasets spanning SD to FHD (2K) show that \emph{ReDT} outperforms state-of-the-art visual geometry transformers in both reconstruction quality and computational efficiency.
For instance, on the ETH3D (2K) dataset, \emph{ReDT} achieves a $16\times$ speedup, a $34\%$ reduction in peak GPU memory usage, and a $33\%$ reduction in chamfer distance over VGGT-$\Omega$.
\href{https://redt.github.io}{Project Website.}
\end{abstract}

%% file: Sections/Introduction.tex
\section{Introduction}

Multi-view reconstruction is a long-standing problem in computer vision, with broad applications in virtual reality~\citep{wang2026samt}, robotics~\citep{han2026geometric}, and autonomous driving~\citep{xiong2026vggtmotion}. 
Traditional approaches, such as Structure-from-Motion (SfM)~\citep{schoenberger2016sfm} and Multi-View Stereo (MVS)~\citep{schoenberger2016mvs}, address this problem through iterative pipelines that scale poorly with the number of input views.
Recently, end-to-end approaches such as VGGT and its successors have demonstrated the effectiveness of feed-forward models for reconstructing 3D structures in a single forward pass~\citep{wang2025vggt,wang2026vggtomega,depthanything3,wang2026pi,keetha2026mapanything}. 
Recent efficient variants have further improved the scalability of these models with respect to view count by reducing the complexity of inter-view (global) attention from quadratic to linear using test-time training (TTT)~\citep{elflein2026vggttt,jin2026zipmap}.

Despite these advances, a critical limitation persists: \textit{existing visual geometry transformers suffer significant degradation in reconstruction quality on HD inputs, while their computational cost grows quadratically with input resolution} (Fig.~\ref{fig:motiv_A}).
We argue that the bottleneck lies in the attention mechanism, which becomes increasingly inefficient and ineffective as input resolution grows.
In particular, existing models mostly employ scaled dot-product attention (SDPA), whose computational cost grows quadratically with the number of tokens, making it increasingly expensive at higher input resolutions.
Moreover, our experiments show that SDPA in visual geometry transformers learns spatial relationships that depend on the training resolution, limiting generalization to higher resolutions (Fig.~\ref{fig:motiv_A}c).
Consequently, they not only become substantially slower as input resolution increases, but also suffer from degraded reconstruction quality at high resolutions, failing to fully exploit the richer spatial information provided by higher-resolution inputs.

To address this limitation, we introduce \emph{ReDT}, a feed-forward visual geometry transformer designed to scale multi-view 3D reconstruction to high-resolution inputs efficiently and effectively.
Intuitively, higher-resolution inputs provide finer spatial detail that can improve reconstruction quality, but naively increasing the input resolution can lead to prohibitive computational costs due to the quadratic growth of standard attention mechanisms.
Through our empirical analysis (Fig.~\ref{fig:motiv_B}), we observe that intermediate per-view representations in visual geometry transformers exhibit a nearly constant effective rank as both input resolution and view count increase.
This motivates us to use a fixed-size set of latent basis tokens per view to efficiently capture high-resolution spatial information without incurring the quadratic computational cost of SDPA.
To this end, \emph{ReDT} replaces the self-attention blocks in the encoder and aggregator with our Discretization-Invariant Visual Attention (DIVA) blocks.
Specifically, DIVA projects patch-wise image features onto a compact set of basis tokens using a density-aware learned projection.
Cross-view correspondences are then computed over this resolution-stable latent representation using test-time training before the resulting features are projected back to the patch domain for decoding.

This design provides two key advantages.
First, \emph{ReDT} is substantially more robust to changes in input resolution than existing 3D reconstruction models. 
We attribute this robustness to its attention formulation, which uses quadrature-aware basis projections to approximate the same continuous operator across different input resolutions. 
This enables \emph{ReDT} to be trained at lower resolutions and subsequently scaled to higher-resolution inputs without additional training. 
Second, the size of its intermediate representation is independent of input resolution, avoiding the quadratic growth in attention cost that typically accompanies higher-resolution inputs (as illustrated in Fig.~\ref{fig:main}). 
This allows \emph{ReDT} to process 100 frames up to FHD (2K) input resolutions with a maximum GPU memory of 61 GB and processing time of $8.2\,$s.
Through extensive experiments across nine datasets, we show that \emph{ReDT} achieves state-of-the-art performance on both HD and conventional SD benchmarks.

In summary, our \textbf{contributions} are as follows:
\begin{itemize}
\item We systematically study the effect of resolution on visual geometry transformers and show that they suffer from performance degradations and poor  scaling at higher resolutions.

\item We introduce \emph{ReDT}, a resolution-aware visual geometry transformer informed by effective rank analysis. Inspired by our effective rank analysis, \emph{ReDT} maps variable-resolution image features into a fixed-size basis tokens for resolution-independent geometric reasoning.

\item We demonstrate across extensive experiments that \emph{ReDT} maintains or improves reconstruction quality as resolution increases, while providing favorable memory usage and runtime.
\end{itemize}

%% file: Sections/Related_work.tex
\section{Related Work}
\subsection{Multi-View Reconstruction}
Reconstructing 3D scene geometry from image collections is a long-standing problem in computer vision.
Recent work has introduced feed-forward models that directly infer 3D geometry from image inputs.
DUSt3R~\citep{dust3r_cvpr24} predicts dense point maps from image pairs in a feed-forward manner, while VGGT~\citep{wang2025vggt} extends this paradigm to unordered multi-view inputs using a Visual Geometry Grounded Transformer.
Subsequent methods, including DepthAnything3~\citep{depthanything3}, $\pi^3$~\citep{wang2026pi}, MapAnything~\citep{keetha2026mapanything}, and VGGT-$\Omega$~\citep{wang2026vggtomega}, further advance feed-forward multi-view 3D reconstruction with improved reconstruction quality and generalization.
Despite this progress, visual geometry transformers are still predominantly trained and evaluated at fixed, relatively low resolutions, and become increasingly expensive as image resolution grows.
This is largely due to the self-attention layers used in the image encoder and alternating attention blocks, whose cost scales quadratically with token count.
More recently, HD-VGGT~\citep{chen2026hd} attempts to address the challenge of high-resolution inputs by selectively extracting high-resolution features, but it still incurs substantial computational cost, does not fully resolve the scalability issue, and is evaluated mostly under standard-resolution settings.
In contrast, \emph{ReDT} reduces attention complexity to linear in the number of views and near-constant with respect to input resolution, while improving reconstruction quality for both SD and HD inputs under comprehensive evaluation settings.

\subsection{Efficient Multi-View Reconstruction}
A major limitation of the conventional visual geometry transformers is the quadratic cost of attention with respect to token count.
Recent efforts address this limitation along two distinct directions.
The first direction focuses on reducing the number of tokens processed by attention.
This can be achieved by merging redundant tokens~\citep{shen2026fastvggt}, reducing the key-value set ~\citep{wang2026flashvggt}, learning compact representative tokens~\citep{huang2026turbovggt}, etc.
A second line of work has instead modified the attention mechanism itself, replacing quadratic attention with more efficient variants, such as the test-time training (TTT) used by VGGT3~\citep{elflein2026vggttt} and ZipMap~\citep{jin2026zipmap}, which scale linearly with token count.
While these approaches substantially improve efficiency, they mostly target global attention, leaving other resolution-dependent self-attention mechanisms, such as frame attention and encoder attention, unchanged.
As a result, they scale efficiently with the number of views but not with input resolution.
In contrast, \emph{ReDT} targets scalability in terms of both view count and view resolution by combining architectural acceleration with token reduction.
\emph{ReDT} pairs TTT-based global reasoning with compression into a fixed-size set of basis tokens per view, reducing the number of tokens processed throughout the network while maintaining linear complexity with respect to the input token count.
This design lowers the cost of resolution-dependent self-attention while maintaining or improving reconstruction quality across both SD and HD inputs.

\subsection{Resolution-Aware Attention}
Standard Transformers typically operate on a fixed discretization and therefore generalize poorly when the sampling resolution at test time differs from that used during training.
Recent work addresses this limitation through discretization-invariant attention operators~\citep{calvello2025continuum,hao2024dpot,xiao2026functional,yang2026simple,kamkari2026learning}, which model mappings between continuous functions rather than mappings tied to specific sample points.
More efficient variants use either fixed bases, such as Fourier modes~\citep{hao2024dpot}, or learned latent functional representations, as in Functional Attention~\citep{xiao2026functional} and LRSA~\citep{yang2026simple}.
However, directly applying these techniques to visual geometry tasks is non-trivial due to the domain mismatch between 2D images and 3D representations where 3D representations need to capture 3D structure that is not explicitly present in the 2D input.
\emph{ReDT} shares the merit of discretization-invariant attention and proposes a quadrature-aware basis projection that maps resolution-dependent image features to a fixed set of discretization-invariant basis tokens and subsequently performs attention in a discretization-invariant space, which allows for both better scaling and improved performance on larger input sets.

%% file: Sections/Method.tex
\begin{figure}[t]
\centering
\includegraphics[width=0.99\textwidth]{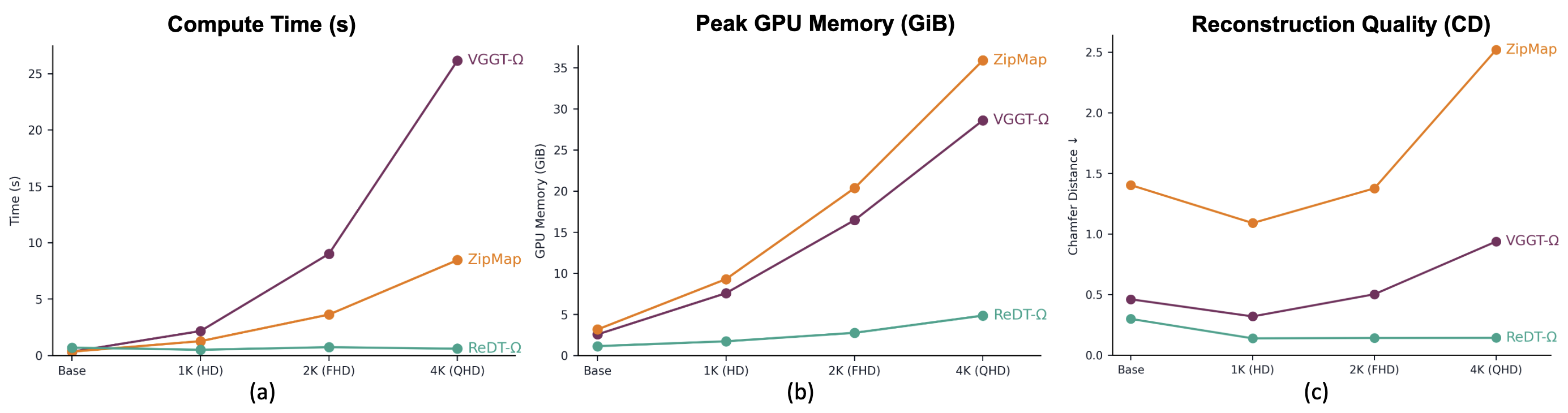}
\caption{Reconstruction performance on the 20-view ETH3D (4K) dataset across input resolutions. (a-b) Runtime and peak GPU memory without task heads. (c) Reconstruction quality (CD).
Pixel dimensions for each resolution are provided in the appendix.}
\label{fig:motiv_A}
\end{figure}

\section{ReDT: resolution-Aware Visual Geometry Transformer}

\subsection{Preliminaries: Visual Geometry Transformers}

We begin with a brief overview of visual geometry transformer architectures.
\emph{ReDT} builds on the VGGT-family models~\citep{wang2025vggt,wang2026vggtomega}; therefore, we focus our discussion on their underlying architecture.
VGGT~\citep{wang2025vggt} takes as input an ordered collection of 2D images and encodes each view using a DINO image encoder.
The resulting patch tokens are augmented with a camera token and a set of register tokens, and are then processed by a sequence of alternating attention blocks.
Alternating attention consists of 24 layers of repeating intra-frame and inter-frame attention.
Intra-frame attention (Frame Attention) operates independently over the tokens within each view, while inter-frame attention (Global Attention) enables information exchange across tokens from all views.
Finally, the resulting representations are passed to dense-prediction and camera heads to predict 3D scene geometry.
VGGT-$\Omega$~\citep{wang2026vggtomega} is an improved variant of VGGT that simplifies the original architecture and enhances its efficiency and scalability.
In particular, it introduces register attention to enable efficient inter-frame information exchange through a compact set of register tokens, while adopting a unified dense prediction head.
In this work, we primarily present empirical analysis and develop architectural improvements based on VGGT-$\Omega$, while the proposed approach is generally applicable to the broader VGGT-family models.

\subsection{Analysis of Scalability Bottlenecks and Resolution Sensitivity}

\paragraph{Scalability Bottlenecks With Visual Geometry Transformers.}

Feed-forward visual geometry transformers, such as VGGT, rely on an alternating attention mechanism that supports cross-view reasoning with Global Attention and within-view refinement with Frame Attention. 
While effective, this design introduces a large number of self-attention blocks, each of which scales quadratically with input resolution, making inference prohibitively slow and memory-intensive for HD inputs.
As shown in Fig.~\ref{fig:motiv_A}(a--b), increasing the input resolution from HD (1K) to QHD (4K) leads to a quadratic increase in compute.
Even more efficient models that replace Global Attention with TTT, such as ZipMap, still exhibit quadratic growth in runtime and memory consumption, as their image encoders and Frame Attention blocks continue to rely on scaled dot-product attention (SDPA).
Moreover, these attention mechanisms are not resolution-aware, making current visual geometry transformers inherently sensitive to changes in view resolution.
As shown in Fig.~\ref{fig:motiv_A}(c), although increasing the resolution beyond HD (1K) provides richer spatial information and would ideally maintain or improve reconstruction performance, the performance of VGGT-$\Omega$ and ZipMap instead progressively degrades.
We attribute this behavior to attention dilution, where attention becomes more dispersed across the larger set of candidate matches introduced at higher image resolutions. 
Even if the correct match retains the highest score, many weak matches can collectively receive a substantial share of attention. 
Aggregating information from these less relevant locations can weaken the signal from the correct match and destabilize learned correspondences.
Together, these limitations motivate attention mechanisms that can process higher-resolution inputs efficiently while preserving correspondence and reconstruction quality.

\begin{figure}[t]
\centering
\includegraphics[width=0.99\textwidth]{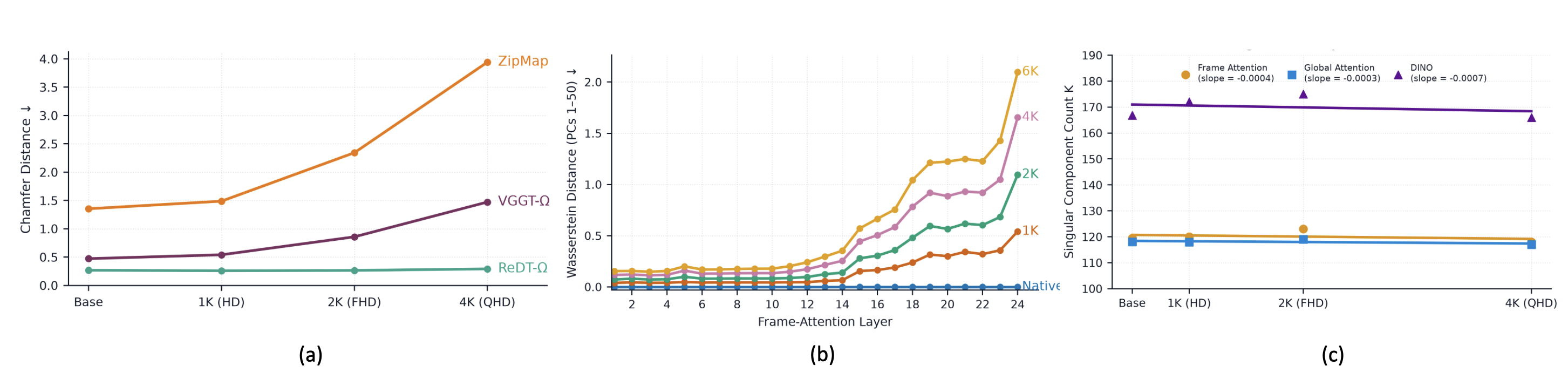}
\caption{All assessments are on the ETH3D dataset. (a) Reconstruction error (CD) under upsampling without added visual information. (b) Mean 1D Wasserstein distance from base-resolution features across VGGT-$\Omega$ frame-attention layers, with curves for different resolutions. (c) Effective rank (99\% energy) of DINO, frame-attention, and global-attention features across resolutions.}
\label{fig:motiv_B}
\end{figure}

\paragraph{Resolution Sensitivity and Effective Rank.}

A natural follow-up to Fig.~\ref{fig:motiv_A}(c) is whether the observed degradation requires additional high-resolution image content.
To investigate this, we downsample each image to the base resolution and then upsample it to 1K, 2K, and 4K.
This changes the image sampling and resulting token representations without introducing additional source detail.
As shown in Fig.~\ref{fig:motiv_B}(a), ZipMap and VGGT-$\Omega$ progressively degrade, closely matching their behavior on native high-resolution inputs.
These results show that changes in image sampling can adversely affect reconstruction even when no additional scene information is available.
We further examine this sensitivity in feature space using the mean one-dimensional Wasserstein distance across principal components relative to the base-resolution features.
Fig.~\ref{fig:motiv_B}(b) shows larger feature-distribution differences at higher resolutions, with greater differences in later VGGT-$\Omega$ aggregator layers.
Together, these observations motivate an architecture that reduces the dependence of geometric reasoning on the spatial discretization of its inputs by providing the models a mechanism to decouple the resolution from the representation.
Fig.~\ref{fig:motiv_B}(c) provides a complementary motivation: \emph{intermediate representations remain compact despite substantial increases in spatial token count.}
We perform per-view SVD on intermediate VGGT-$\Omega$ features and measure the number of components required to explain $99\%$ of their energy.
Across the examined datasets, resolutions, and view counts, this effective rank remains nearly constant for DINO, frame-attention, and global-attention features.
This empirical regularity motivates a fixed-size latent representation for feature interactions, while retaining spatial representations for dense prediction.

\paragraph{Summary.}
Our analysis identifies increasing computational cost and sensitivity to changes in image sampling as two challenges when scaling visual geometry transformers to high-resolution inputs.
Meanwhile, the approximately stable effective rank of intermediate features suggests that compact representations can support efficient feature interactions.
These findings motivate ReDT’s separation of spatial token count from the per-view latent interaction budget.
Its area-aware aggregation additionally accounts for the unequal spatial support of tokens produced by adaptive patch merging.

\subsection{ReDT Model Architecture}

\begin{figure}[t]
\centering
\includegraphics[width=0.99\textwidth]{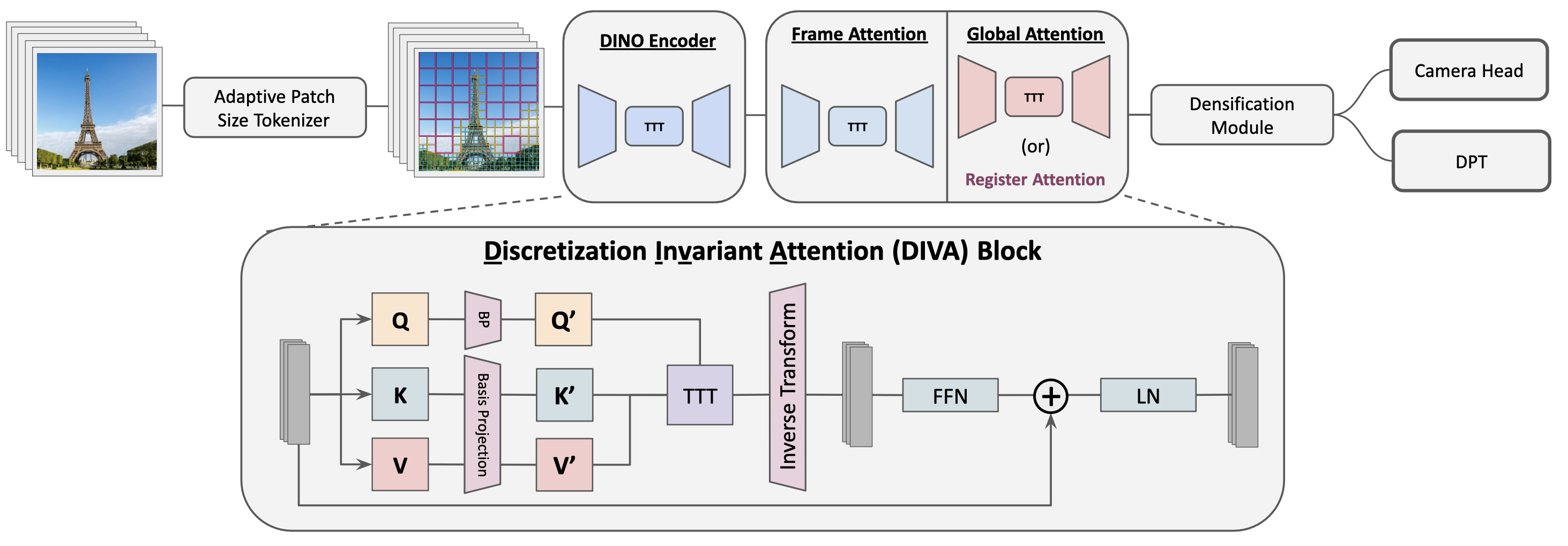}
\caption{ReDT architecture with adaptive tokenization, DIVA blocks, and dense prediction heads.}
\label{fig:model_arch}
\end{figure}

\paragraph{Overview.}
We introduce \emph{ReDT}, a resolution-aware visual geometry transformer for efficient HD reconstruction.
We begin by summarizing its three key architectural changes over existing visual geometry transformers (see Fig.~\ref{fig:model_arch}).
First, ReDT uses a variable-patch-size encoder that adaptively merges nearby low-entropy patches in a quadtree-like manner, reducing spatial redundancy and producing a compact set of variable-size tokens.
Second, it replaces standard VGGT-$\Omega$ attention with our discretization invariant attention (DIVA) blocks, which compresses features into \(K\) input-dependent basis tokens, performs correspondence reasoning in this compact space, and projects them back to the original resolution.
Finally, DIVA replaces the remaining correspondence mechanisms in the encoder and aggregator with test-time training (TTT).
Together, these changes improve robustness to resolution shifts while substantially reducing memory and runtime.
As illustrated in Fig.~\ref{fig:motiv_A}(a--b), the run-time complexity of \emph{ReDT} remains nearly constant as input resolution increases.
Peak memory grows slightly because higher-resolution images must still occupy memory, while runtime remains approximately constant.
Moreover, as shown in Fig.~\ref{fig:motiv_A}(c), \emph{ReDT} preserves the strong reconstruction performance achieved at 1K resolution even as the input resolution is increased.

\paragraph{Variable Patch Size Encoder.}
ReDT employs a variable-patch-size tokenizer and encoder to reduce spatial token count while preserving fine-grained information.
Given an input image \(I\), we first extract features on the encoder’s native patch grid and compute an image-space entropy score for each patch.
A quadtree then adaptively determines the spatial resolution by recursively merging neighboring low-entropy patches into larger \(2\times\), \(4\times\), and \(8\times\) patch regions, while high-entropy regions retain the native patch resolution.
The corresponding features are aggregated according to this partition, producing a variable-length token sequence that allocates more tokens to information-dense regions and fewer to locally redundant regions before ViT encoding.

\paragraph{Discretization Invariant Attention Blocks.}
ReDT replaces all attention blocks within the VGGT-$\Omega$'s image encoder and backbone with DIVA blocks.
We adopt the term discretization-invariant following established usage in the literature~\citep{calvello2025continuum,xiao2026functional,yang2026simple}.
For attention formulations such as DIVA, the intended notion is asymptotic discretization consistency, which does not require exact equality across finite resolutions.
At each transformer layer and attention head, with the exception of register-attention blocks, DIVA constructs token-to-basis assignment operators that map \(N\) spatial tokens to \(K\) latent basis vectors. Specifically, following the query and key/value projections ($f_Q$ and $f_{KV}$), we predict assignment matrices \(A_Q,A_{KV}\in\mathbb{R}^{N\times K}\):
\[
A^Q = \operatorname{softmax}_{K}\!\left(f_Q(Q)\right),
\qquad
A^{KV} = \operatorname{softmax}_{K}\!\left(f_{KV}(K)\right).
\]
The query, key, and value features are then projected into the latent space using quadrature weights \(c_i\), which represent the spatial area associated with token \(i\) according to its patch size as follows:
\[
\widetilde{Q}_k
=
\frac{\sum_{i=1}^{N} c_i A^{Q}_{ik} Q_i}
{\sum_{i=1}^{N} c_i A^{Q}_{ik}+\epsilon},
\qquad
\widetilde{K}_k
=
\frac{\sum_{i=1}^{N} c_i A^{KV}_{ik} K_i}
{\sum_{i=1}^{N} c_i A^{KV}_{ik}+\epsilon},
\qquad
\widetilde{V}_k
=
\frac{\sum_{i=1}^{N} c_i A^{KV}_{ik} V_i}
{\sum_{i=1}^{N} c_i A^{KV}_{ik}+\epsilon}.
\]
Correspondences are computed entirely in the \(K\)-token latent space using the TTT procedure introduced in ZipMap~\citep{jin2026zipmap}.
The resulting latent features are mapped back to the original \(N\)-token spatial discretization using the transpose of the spatial-to-latent assignment operator, which empirically achieves reconstruction quality comparable to the pseudo-inverse.
In our DIVA blocks, special tokens, including CLS, register, and camera tokens, are excluded from area-weighted compression, as they do not correspond to a well-defined spatial measure, and are instead concatenated with the compressed basis tokens.

\paragraph{ReDT Decoder/Task Heads.}
Following VGGT-$\Omega$, we cache features from layers 5, 12, 18, and 24 and pass them to a DPT decoder, while camera tokens are passed to a token-based camera head.
We leave these downstream heads architecturally unchanged, as our focus is on the learned representations themselves.
Since the heads primarily serve to decode latent features into task-specific outputs, they can in principle be replaced by alternative formulations.
Before passing the features to DPT, we densify the quadtree representation by broadcasting each token over its corresponding spatial region, thereby discarding the explicit quadtree structure, and apply a convolutional layer to smooth the resulting feature map and introduce local spatial variation that DPT can exploit for recovering low-amplitude details.

%% file: Sections/Experiments.tex
\section{Experiments}

We evaluate \emph{ReDT} on point-cloud reconstruction using summed Chamfer Distance (CD), camera-pose estimation using Relative Translation Accuracy (RTA), Relative Rotation Accuracy (RRA), and AUC@30$^\circ$, and depth estimation using Absolute Relative Error (AbsRel) and threshold accuracy ($\delta < 1.25$), following~\citet{wang2025vggt,shen2026fastvggt,wang2026vggtomega}.
We also evaluate \emph{ReDT}, which applies ReDT to the original VGGT backbone for direct comparison with FastVGGT, FlashVGGT, and ZipMap.
Our HD benchmark includes seven datasets spanning approximately 1K-8K resolution to assess reconstruction quality and scalability.
Given the limited coverage of high-resolution 3D reconstruction in literature, we also evaluate on ScanNetv2 and 7-Scenes for comparison with prior work.
Further experimental details are in the appendix.

\input{tables/point_cloud_HD}
\input{tables/pose_depth_HD}
\input{tables/point_cloud_subHD}

\subsection{High-Definition Reconstruction}
We evaluate \emph{ReDT} across seven datasets spanning diverse scene dynamics and high-resolution inputs.
These include DynamicReplica~\citep{karaev2023dynamicstereo}, SAIL-VOS 3D~\citep{hu2021sailvos3d}, MVS-Synth~\citep{huang2018deepmvs}, ScanNet++~\citep{yeshwanth2023scannetpp}, Spring~\citep{mehl2023spring}, UnrealStudios4K~\citep{tosi2021smdnets}, and ETH3D~\citep{schops2017eth3d}.
ReDT uses the VGGT backbone to assess the benefits of our approach against other VGGT-based methods, while ReDT-$\Omega$ uses a VGGT-$\Omega$ backbone and is our primary model.

Table~\ref{tab:pointcloud_HD_results} shows that foundation models (VGGT, VGGT-$\Omega$, $\pi^3$, DepthAnything3) generally achieve lower Chamfer Distance (CD) than efficient baselines (FastVGGT, FlashVGGT, ZipMap), whereas efficient methods offer shorter runtimes.
With the VGGT backbone, \emph{ReDT} improves upon VGGT and all evaluated efficient baselines in both CD and runtime across all four reconstruction datasets.
These results demonstrate that the benefits of our approach are already evident with the VGGT backbone.
For example, on ScanNet++, \emph{ReDT} reduces CD from 0.6663 for ZipMap to 0.1388, while reducing runtime from 7.84\,s to 5.59\,s.
Our primary model, \emph{ReDT-$\Omega$}, further improves both accuracy and runtime over \emph{ReDT}, achieving the lowest CD and shortest runtime across all four reconstruction datasets.
Among foundation models, VGGT-$\Omega$ achieves the lowest CD on SAIL-VOS 3D, UnrealStudios4K, and ETH3D, while DepthAnything3 leads on ScanNet++.
Relative to the strongest foundation-model baseline on each dataset, \emph{ReDT-$\Omega$} reduces CD by approximately $25.1\%$ while providing up to a $16.17\times$ speedups.
It also runs up to $4.54\times$ faster than ZipMap, the fastest competing efficient method on every dataset, while consistently achieving lower CD.
Thus, \emph{ReDT} establishes the effectiveness of our approach with the VGGT backbone, while \emph{ReDT-$\Omega$} delivers the strongest overall accuracy and runtime.

The pose and depth results in Table~\ref{tab:pose_depth_hd} further support the performance of our primary model.
For camera pose estimation, \emph{ReDT-$\Omega$} improves AUC@30 over the strongest foundation-model baseline by 9.26 percentage points on MVS-Synth and 14.43 percentage points on ScanNet++, while achieving the highest RTA and matching the best RRA on both datasets.
Both \emph{ReDT} variants also surpass all evaluated efficient baselines across the reported pose metrics, demonstrating that the gains extend to the VGGT-based variant.
For depth estimation, \emph{ReDT-$\Omega$} reduces AbsRel relative to the strongest baseline, VGGT-$\Omega$, by 45.9\% on DynamicReplica and 13.3\% on Spring, while achieving the highest $\delta_{1.25}$ accuracy on both datasets.
Together, these results establish \emph{ReDT-$\Omega$} as our strongest evaluated model across the three geometric tasks, with \emph{ReDT} providing a complementary comparison that demonstrates the effectiveness of our approach with the VGGT backbone.

\subsection{Comparison on Standard Sub-HD Benchmarks}
While our primary focus is high-resolution reconstruction, we additionally evaluate both \emph{ReDT} variants on standard sub-HD benchmarks to facilitate comparison with prior work.
Specifically, we assess reconstruction quality and runtime on ScanNet~\citep{dai2017scannet} and 7-Scenes~\citep{shotton2013scene}, noting that ScanNet is distinct from ScanNet++.
Table~\ref{tab:pointcloud_all_results} shows that both variants achieve competitive reconstruction quality with consistently low inference time across sequence lengths and sampling densities.
With the VGGT backbone, \emph{ReDT} preserves reconstruction quality on ScanNet while substantially reducing runtime, achieving speedups of up to $12.7\times$.
It also consistently achieves lower Chamfer Distance (CD) than ZipMap while matching or reducing its runtime.
These results demonstrate that the benefits of our approach extend to standard sub-HD benchmarks with the VGGT backbone, particularly as sequence length increases.
Our primary model, \emph{ReDT-$\Omega$}, improves both reconstruction quality and runtime over its VGGT-$\Omega$ backbone on ScanNet, and consistently outperforms ZipMap in both measures.
On 7-Scenes, both \emph{ReDT} variants achieve the lowest reported CD ($0.032$) under sparse sampling and closely approach the best reconstruction quality under dense sampling, with substantially shorter runtimes than $\pi^3$, the most accurate method in the dense setting.
Although FlashVGGT retains a small runtime advantage on ScanNet-100 and sparse 7-Scenes, both \emph{ReDT} variants are faster on ScanNet-300, ScanNet-500, and dense 7-Scenes.
Together, these results show that our approach remains effective beyond high-resolution reconstruction, combining competitive accuracy with favorable runtime scaling across both backbones.

\input{tables/ablation_component} 

\input{tables/ablation_threshold}

\subsection{Ablation Study}

\paragraph{Component Ablation.}
We evaluate the effectiveness of the adaptive tokenizer, basis projection, and test-time training by removing each component individually from the full \emph{ReDT-$\Omega$} model.
Table~\ref{tab:component_ablation} reports results on ETH3D at 1K, 2K, and 4K resolutions.
The full model achieves the best performance across all metrics and resolutions.
Removing the adaptive tokenizer consistently degrades accuracy, with the largest increase in AbsRel at 4K, from 0.0425 to 0.0943, highlighting its importance for high-resolution inputs.
Removing basis projection produces the largest degradation at 1K and 2K in both CD and AbsRel and leads to an out-of-memory failure at 4K, demonstrating its importance for both accuracy and memory efficiency.
Finally, removing test-time training modestly increases CD and AbsRel but reduces AUC@30 by 4.16-4.52 points across resolutions, indicating a consistent benefit from test-time adaptation.

\paragraph{Basis Count Ablation.}
We investigate the sensitivity of \emph{ReDT} to the number of basis elements in the encoder and aggregator.
Starting from the default configuration of $K_{\mathrm{enc}}=512$ and $K_{\mathrm{agg}}=64$, we vary each count independently while keeping the other fixed.
For the encoder, we reduce $K_{\mathrm{enc}}$ to 256 to assess whether a more compact basis preserves reconstruction quality.
For the aggregator, we increase $K_{\mathrm{agg}}$ to 128 to evaluate whether additional basis capacity improves performance.
As shown in Table~\ref{tab:ablation_basis_aggregator}, reducing the encoder basis count increases CD from 0.1137 to 0.3737 and decreases AUC@30 from 86.09 to 36.18, while AbsRel increases slightly from 0.0356 to 0.0367.
Increasing the aggregator basis count also degrades all three metrics, yielding a CD of 0.1420, an AbsRel of 0.0430, and an AUC@30 of 81.79.
These results support the default configuration, which achieves the best performance across all three metrics among the tested settings, and indicate that increasing aggregator capacity alone does not necessarily improve reconstruction quality.

%% file: tables/point_cloud_HD.tex
\begin{table*}[t]
\centering
\caption{
Point-cloud reconstruction comparison across four high-resolution datasets.
Lower Chamfer Distance (CD), runtime, and peak memory are better.
Best results are \textbf{bold}, second-best results are \underline{underlined}, and third-best
results are \underline{\underline{double-underlined}}.
}
\label{tab:pointcloud_HD_results}

\setlength{\tabcolsep}{4pt}
\renewcommand{\arraystretch}{1.15}

\resizebox{0.95\textwidth}{!}{%
\begin{tabular}{l|ccc|ccc|ccc|ccc}
\toprule

\multirow{2}{*}{\textbf{Method}}
& \multicolumn{3}{c|}{\textbf{SAIL-VOS 3D (1K)}}
& \multicolumn{3}{c|}{\textbf{ScanNet++ (1K)}}
& \multicolumn{3}{c|}{\textbf{UnrealStudios4K (2K)}}
& \multicolumn{3}{c}{\textbf{ETH3D (2K)}} \\

\cmidrule(lr){2-4}
\cmidrule(lr){5-7}
\cmidrule(lr){8-10}
\cmidrule(lr){11-13}

&
\textbf{CD $\downarrow$} & \textbf{Time (s)} & \textbf{Mem. (GB)} &
\textbf{CD $\downarrow$} & \textbf{Time (s)} & \textbf{Mem. (GB)} &
\textbf{CD $\downarrow$} & \textbf{Time (s)} & \textbf{Mem. (GB)} &
\textbf{CD $\downarrow$} & \textbf{Time (s)} & \textbf{Mem. (GB)} \\

\midrule

VGGT (CVPR 2025)
& 23.0525 & 59.00 & 47.25
& 1.0481 & 58.90 & 43.25
& 9.8623 & 116.88 & 101.72
& 2.5660 & 118.02 & 102.24 \\

VGGT-$\Omega$ (CVPR 2026)
& \underline{11.8681} & 28.67 & \underline{\underline{36.77}}
& \underline{\underline{0.1362}} & 28.67 & 38.42
& \underline{1.0299} & 102.66 & \underline{\underline{93.24}}
& \underline{0.1702} & 102.70 & \underline{\underline{98.09}} \\

$\pi^3$ (ICLR 2026)
& 12.7053 & 55.52 & 37.94
& 0.3649 & 55.45 & \underline{\underline{33.56}}
& 3.5208 & 109.46 & 98.57
& 0.7768 & 108.79 & 101.54\\

DepthAnything3 (ICLR 2026)
& 14.3069 & 20.90 & 46.77
& \underline{0.1081} & 20.89 & 39.44
& 2.5909 & 80.91 & 97.82
& \underline{\underline{0.3158}} & 80.88 & 102.80 \\

\midrule

FastVGGT (ICLR 2026)
& 23.0125 & 21.32 & 34.11
& 0.8946 & 21.31 & 28.77
& 8.3933 & 84.29 & 94.88
& 2.4710 & 84.30 & 98.51 \\

FlashVGGT (CVPR 2026)
& 32.0492 & 9.15 & 40.06
& 0.9762 & 9.13 & 33.78
& 8.4182 & 36.63 & 100.96
& 2.7161 & 36.46 & 105.22 \\

ZipMap (CVPR 2026)
& 13.9967 & \underline{\underline{7.83}} & 48.93
& 0.6663 & \underline{\underline{7.84}} & 44.18
& 9.3427 & \underline{\underline{28.84}} & 119.82
& 1.7539 & \underline{\underline{28.83}} & 121.56 \\

\midrule

\textbf{ReDT (Ours)}
& \underline{\underline{12.6891}} & \underline{6.27} & \underline{36.38}
& 0.1388 & \underline{5.59} & \underline{30.68}
& \underline{\underline{1.5524}} & \underline{12.21} & \underline{67.20}
& 0.5476 & \underline{12.49} & \underline{71.07} \\

\textbf{ReDT-$\Omega$ (Ours)}
& \textbf{11.0931} & \textbf{3.05} & \textbf{30.98}
& \textbf{0.0501} & \textbf{3.16} & \textbf{25.68}
& \textbf{0.9575} & \textbf{6.84} & \textbf{61.63}
& \textbf{0.1140} & \textbf{6.35} & \textbf{64.47} \\

\bottomrule
\end{tabular}
}
\end{table*}

%% file: tables/pose_depth_HD.tex
\begin{table*}[t]
\centering
\caption{
Camera pose estimation on MVS-Synth (2K) and ScanNet++ (2K), and depth estimation on DynamicReplica (1K) and Spring (2K).
All methods use the fixed-HD evaluation protocol.
}
\label{tab:pose_depth_hd}

\setlength{\tabcolsep}{4pt}
\renewcommand{\arraystretch}{1.15}

\resizebox{0.95\textwidth}{!}{%
\begin{tabular}{l|cccccccccc}
    \toprule

    \multirow{3}{*}{\textbf{Method}}
    & \multicolumn{6}{c}{\textbf{Camera Pose Estimation}}
    & \multicolumn{4}{c}{\textbf{Depth Estimation}} \\

    \cmidrule(lr){2-7}
    \cmidrule(lr){8-11}

    & \multicolumn{3}{c}{\textbf{MVS-Synth (2K)}}
    & \multicolumn{3}{c}{\textbf{ScanNet++ (2K)}}
    & \multicolumn{2}{c}{\textbf{DynamicReplica (1K)}}
    & \multicolumn{2}{c}{\textbf{Spring (2K)}} \\

    \cmidrule(lr){2-4}
    \cmidrule(lr){5-7}
    \cmidrule(lr){8-9}
    \cmidrule(lr){10-11}

    & \textbf{AUC@30 $\uparrow$}
    & \textbf{RTA $\uparrow$}
    & \textbf{RRA $\uparrow$}
    & \textbf{AUC@30 $\uparrow$}
    & \textbf{RTA $\uparrow$}
    & \textbf{RRA $\uparrow$}
    & \textbf{AbsRel $\downarrow$}
    & $\boldsymbol{\delta < 1.25}\ \boldsymbol{\uparrow}$
    & \textbf{AbsRel $\downarrow$}
    & $\boldsymbol{\delta < 1.25}\ \boldsymbol{\uparrow}$ \\

    \midrule

    VGGT
    & 14.16 & 40.11 & 88.21
    & 5.04 & 36.80 & 44.74
    & 0.1748 & 0.7812
    & 0.3856 & 0.4473 \\

    VGGT-$\Omega$
    & 83.46 & 98.80 & {100.00}
    & 69.99 & 93.33 & {100.00}
    & 0.0246 & 0.9938
    & 0.0692 & 0.9457 \\

    $\pi^3$
    & 20.83 & 44.58 & {100.00}
    & 44.98 & 79.70 & 95.67
    & 0.0655 & 0.9702
    & 0.2586 & 0.5586 \\

    DepthAnything3
    & 75.28 & 94.30 & {100.00}
    & 79.22 & 98.28 & 99.65
    & 0.0260 & 0.9944
    & 0.1638 & 0.7267 \\

    \midrule

    FastVGGT
    & 14.57 & 39.62 & 89.91
    & 5.74 & 38.09 & 44.31
    & 0.1918 & 0.7400
    & 0.3883 & 0.4442 \\

    FlashVGGT
    & 13.29 & 39.98 & 85.38
    & 6.17 & 36.94 & 50.69
    & 0.1567 & 0.8524
    & 0.3903 & 0.4308 \\

    ZipMap
    & 19.38 & 49.82 & 94.90
    & 24.03 & 63.09 & 91.83
    & 0.1096 & 0.9337
    & 0.1701 & 0.7379 \\

    \midrule

    \textbf{ReDT (Ours)}
    & 53.46 & 80.20 & {100.00}
    & 69.33 & 94.33 & 99.99
    & 0.0248 & 0.9947
    & 0.1749 & 0.6278 \\

    \textbf{ReDT-$\Omega$ (Ours)}
    & \textbf{92.72} & \textbf{99.71} & \textbf{100.00}
    & \textbf{93.65} & \textbf{99.67} & \textbf{100.00}
    & \textbf{0.0133} & \textbf{0.9959}
    & \textbf{0.0600} & \textbf{0.9661} \\

    \bottomrule
\end{tabular}%
}
\end{table*}

%% file: tables/point_cloud_subHD.tex
\begin{table*}[t]
\centering
\caption{
Point-cloud reconstruction comparison across standard-definition datasets.
Lower CD and time are better. For time, best values are bold, second-best
values are underlined, and third-best values are double-underlined.
}
\label{tab:pointcloud_all_results}

\setlength{\tabcolsep}{5pt}
\renewcommand{\arraystretch}{1.15}

\resizebox{0.95\textwidth}{!}{%
\begin{tabular}{l|cc|cc|cc|cc|cc}
\toprule

\multirow{2}{*}{\textbf{Method}}
& \multicolumn{2}{c|}{\textbf{ScanNet-100 Frames}}
& \multicolumn{2}{c|}{\textbf{ScanNet-300 Frames}}
& \multicolumn{2}{c|}{\textbf{ScanNet-500 Frames}}
& \multicolumn{2}{c|}{\textbf{7-Scenes (Sparse)}}
& \multicolumn{2}{c}{\textbf{7-Scenes (Dense)}} \\

\cmidrule(lr){2-3}
\cmidrule(lr){4-5}
\cmidrule(lr){6-7}
\cmidrule(lr){8-9}
\cmidrule(lr){10-11}

&
\textbf{CD $\downarrow$} & \textbf{Time (s)} &
\textbf{CD $\downarrow$} & \textbf{Time (s)} &
\textbf{CD $\downarrow$} & \textbf{Time (s)} &
\textbf{CD $\downarrow$} & \textbf{Time (s)} &
\textbf{CD $\downarrow$} & \textbf{Time (s)} \\

\midrule

VGGT
& 0.421 & 4.9
& 0.429 & 33.6
& 0.442 & 90.1
& 0.118 & 4.5
& 0.115 & 38.1 \\

VGGT-$\Omega$
& 0.500 & 2.0
& 0.531 & 13.0
& 0.548 & 33.7
& 0.034 & 1.9
& 0.033 & 14.6 \\

$\pi^3$
& 0.471 & 3.6
& 0.480 & 25.5
& 0.479 & 67.4
& 0.032 & \underline{\underline{1.6}}
& 0.033 & 28.7 \\

DepthAnything3
& 0.461 & \underline{\underline{1.6}}
& 0.516 & 9.9
& 0.524 & 25.1
& 0.065 & \underline{1.5}
& 0.064 & 11.0 \\

\midrule

FastVGGT
& 0.416 & 2.8
& 0.430 & 12.3
& 0.437 & 28.4
& 0.109 & 2.7
& 0.115 & 14.2 \\

FlashVGGT
& 0.454 & \textbf{1.3}
& 0.459 & 5.0
& 0.464 & 10.2
& 0.035 & \textbf{1.3}
& 0.033 & 5.7 \\

ZipMap
& 0.465 & \underline{\underline{1.6}}
& 0.474 & \underline{\underline{4.6}}
& 0.483 & \underline{\underline{7.7}}
& 0.059 & \underline{1.5}
& 0.057 & \underline{\underline{4.9}} \\

\midrule

\textbf{ReDT (Ours)}
& 0.420 & \underline{\underline{1.6}}
& 0.428 & \underline{4.3}
& 0.446 & \underline{7.1}
& 0.033 & \underline{1.5}
& 0.032 & \underline{4.5} \\

\textbf{ReDT-$\Omega$ (Ours)}
& 0.416 & \underline{1.5}
& 0.431 & \textbf{4.2}
& 0.451 & \textbf{6.7}
& 0.033 & \underline{1.5}
& 0.032 & \textbf{4.4} \\

\bottomrule
\end{tabular}%
}
\end{table*}

%% file: tables/ablation_component.tex
\begin{table*}[t]
\centering
\caption{
Component-ablation for ReDT on ETH3D at
three resolutions.
Lower CD and AbsRel are better; higher AUC@30 is better.
'--' refers to removing the component.
}
\label{tab:component_ablation}
\setlength{\tabcolsep}{4pt}
\renewcommand{\arraystretch}{1.15}

\resizebox{0.95\textwidth}{!}{%
\begin{tabular}{l|ccc|ccc|ccc}
    \toprule
    \multirow{2}{*}{\textbf{Method}}
    & \multicolumn{3}{c|}{\textbf{ETH3D (1K)}}
    & \multicolumn{3}{c|}{\textbf{ETH3D (2K)}}
    & \multicolumn{3}{c}{\textbf{ETH3D (4K)}} \\
    \cmidrule(lr){2-4}
    \cmidrule(lr){5-7}
    \cmidrule(lr){8-10}
    & \textbf{CD $\downarrow$}
    & \textbf{AbsRel $\downarrow$}
    & \textbf{AUC@30 $\uparrow$}
    & \textbf{CD $\downarrow$}
    & \textbf{AbsRel $\downarrow$}
    & \textbf{AUC@30 $\uparrow$}
    & \textbf{CD $\downarrow$}
    & \textbf{AbsRel $\downarrow$}
    & \textbf{AUC@30 $\uparrow$} \\
    \midrule
    \textbf{ReDT-$\Omega$ (Full)}
    & 0.1137 & 0.0356 & 86.09
    & 0.1140 & 0.0332 & 86.82
    & 0.1249 & 0.0425 & 85.75 \\

    \quad $-$ Adaptive Tokenizer
    & 0.1239 & 0.0644 & 84.16
    & 0.1241 & 0.0627 & 84.74
    & 0.1455 & 0.0943 & 83.77 \\

    \quad $-$ Basis Projection
    & 0.1344 & 0.0859 & 82.99
    & 0.1631 & 0.1236 & 78.87
    & OOM & OOM & OOM \\

    \quad $-$ Test-Time Training
    & 0.1195 & 0.0373 & 81.93
    & 0.1198 & 0.0348 & 82.30
    & 0.1313 & 0.0447 & 81.50 \\
    
    \bottomrule
\end{tabular}%
}
\end{table*}

%% file: tables/ablation_threshold.tex
\begin{table}[t]
\centering
\caption{
Ablation of encoder and aggregator basis counts.
}
\label{tab:ablation_basis_aggregator}
\setlength{\tabcolsep}{5pt}
\renewcommand{\arraystretch}{1.15}
\resizebox{0.4\textwidth}{!}{%
\begin{tabular}{ccccc}
\toprule
$K_{\mathrm{enc}}$ &
$K_{\mathrm{agg}}$ &
\textbf{CD $\downarrow$} &
\textbf{AbsRel $\downarrow$} &
\textbf{AUC@30 $\uparrow$} \\
\midrule
512 & 64 & 0.1137 & 0.0356 & 86.09  \\
\midrule
256 & 64  & 0.3737 & 0.0367 & 36.18 \\
512 & 128 & 0.1420 & 0.0430 & 81.79 \\ 
\bottomrule
\end{tabular}%
}
\end{table}

%% file: Sections/Conclusion.tex
\section{Conclusion}

In this work, we propose ReDT for efficient and robust multi-view 3D reconstruction from high-definition inputs. Our approach is motivated by VGGT’s poor scaling with input resolution and the nearly constant effective rank of its intermediate representations. ReDT exploits this structure by projecting backbone features into a fixed-size basis space for efficient geometric reasoning. Attention is then computed through test-time training in this compact space before the outputs are mapped back to the full token space. Extensive experiments on standard-definition (SD) and high-definition (HD) benchmarks demonstrate the effectiveness of our approach. The VGGT-based ReDT variant improves reconstruction accuracy and efficiency over competing VGGT-based methods, while our primary model, ReDT-$\Omega$, achieves the best overall HD reconstruction quality and lowest computational cost among evaluated methods, alongside competitive reconstruction quality and efficiency on SD inputs.

%% file: Sections/Appendix.tex
\section{Appendix}

\subsection{Implementation Details}

We use a patch size of \(16\), with descriptor budgets of \(512\) and \(64\) tokens for the image encoder and global aggregation module, respectively. The camera and depth prediction heads are enabled, while the alignment head is disabled. All model computation is performed in BF16. For global test-time training (TTT), we perform five internal Muon optimization steps during each forward pass using a TTT learning rate of \(0.01\).

We fine-tune all benchmarked models from their publicly available pretrained checkpoints within a unified multi-view geometry training framework, using a mixture of synthetic and real-world datasets. Training follows a mixed-resolution, mixed-view sampling strategy, with each distributed batch drawn from either high-definition (HD) inputs with 2–4 views or standard-definition (SD) inputs with 10–20 views. The number of input views and image aspect ratio are randomly sampled within the selected regime. Images are resized to the target resolution, with camera intrinsics adjusted accordingly, and standard photometric augmentations are applied. This strategy exposes the models to both fine spatial detail and longer multi-view sequences, supporting generalization to HD inputs while maintaining long-view reasoning.

All methods are trained for 10 epochs using AdamW with a base learning rate of \(1\times10^{-5}\), \(\beta=(0.9,0.95)\), and a weight decay of \(0.05\). The pretrained image encoder uses a reduced learning rate of \(5\times10^{-7}\). Learning rates increase linearly during the first epoch and subsequently decay according to a cosine schedule. Gradients are accumulated over six micro-batches and clipped to a global norm of \(1.0\) before each optimizer update. Training uses distributed data parallelism and BF16 mixed precision, with activation checkpointing enabled where applicable to reduce memory consumption.

Predictions and ground truth are expressed relative to the first input camera and independently normalized by their respective multi-view scene scales, making the training objective largely invariant to the global coordinate frame and metric scale. Model forward passes run under autocast, while geometric transformations, normalization, and loss computation are performed in FP32 for numerical stability.

\subsection{Training Losses}
We supervise both global scene geometry and local surface structure during training.
The overall objective is shown below, where $\lambda_o$ controls the strength of the orthonormal regularization:
\begin{equation}
\mathcal{L}
=
\mathcal{L}_{\mathrm{geometry}}
+
\lambda_o \mathcal{L}_{\mathrm{orthonormal}}.
\end{equation}

\paragraph{Geometry Loss.}
The geometry loss jointly supervises the predicted 3D structure, depth, camera pose, and local surface geometry. Specifically, we optimize losses on the predicted 3D points in each camera coordinate system (\texttt{cam\_pts3d}), their $z$-coordinate (\texttt{depth\_z}), relative camera rotation (\texttt{pose\_quats}), and scale-normalized relative camera translation (\texttt{pose\_trans}). For quaternion supervision, we account for the sign ambiguity by taking the minimum error with respect to $q$ and $-q$, since both represent the same rotation.

\paragraph{Surface Geometry Loss.}
We additionally supervise local geometric consistency using normal and depth-gradient losses. The normal loss (\texttt{normal}) measures the angular discrepancy between surface normals constructed from neighboring predicted and ground-truth 3D points. The gradient-matching loss (\texttt{gradient\_matching}) compares log-depth gradients at multiple spatial scales, using strides $1$, $2$, $4$, and $8$, encouraging the model to preserve both fine- and coarse-scale depth structure.

\paragraph{Orthonormal Loss.}
We regularize the assignment matrix \(A\in\mathbb{R}^{N\times K}\) through its Gram matrix,
\[
\mathcal L_{\mathrm{reg}}=\|AA^\top-I_N\|_F^2.
\]
This objective is equivalent, up to the additive constant \(N-K\), to \(\|A^\top A-I_K\|_F^2\), which penalizes correlations between latent assignment columns and deviations of their squared norms from one. We use this objective as a soft regularizer without assuming that the assignments become exactly orthonormal.

\subsection{Training Datasets}

We train on seven datasets spanning real and synthetic scenes
with metric and relative depth. ASE, ParallelDomain4D, and
UnrealStereo4K provide synthetic data with metric depth.
The real datasets include MPSD with metric depth, and
BlendedMVS, DL3DV, and MegaDepth with relative depth.
Each epoch contains 26,250 samples from each dataset except
UnrealStereo4K, which contributes 2,750 samples, for a total
of 160,250 training samples per epoch.

\subsection{Theoretical Complexity}

Let $n$ denote the number of tokens, $D$ the feature dimension,
$h$ the number of heads, $d=D/h$ the per-head dimension,
$k$ the basis size, and $\tau$ the number of TTT steps.
The model comprises a backbone (encoder and alternating-attention
aggregator) and a DPT head.

\paragraph{DIVA block.}
Per head, basis construction and projection cost $\mathcal{O}(ndk)$,
and $\tau$ TTT updates cost $\mathcal{O}(\tau kd^2)$.
Reconstruction,
$Y=\Phi(\widetilde{Q}W^{(\tau)})$,
costs $\mathcal{O}(kd^2+ndk)$.
Summing over heads and including feature projections and the
feed-forward network gives
\begin{equation}
T_{\mathrm{ReDT}}
=
\mathcal{O}\!\left(
nD^2+nDk+(\tau+1)\frac{kD^2}{h}
\right),
\end{equation}
compared with $\mathcal{O}(nD^2+n^2D)$ for standard attention.
Per-group working memory, excluding model parameters and the
outer-training graph, is
\begin{equation}
M_{\mathrm{ReDT}}
=
\mathcal{O}\!\left(
nD+hnk+kD+\frac{D^2}{h}
\right).
\end{equation}

\paragraph{Full model.}
For $N$ views with $P$ patches each and fixed architectural
hyperparameters, replacing all attention outside the DPT head yields
\begin{equation}
T_{\mathrm{backbone}}=\mathcal{O}(NP),
\qquad
T_{\mathrm{full}}=\mathcal{O}(NP+N^2),
\end{equation}
where the quadratic term arises from the unchanged DPT head.

\subsection{Theoretical Discussion on Discretization Invariance}

\paragraph{Definition.}
Discretization invariance means that a learned operator acts
consistently across different samplings of the same continuous
input, such as a 3D scene. Let $u$ denote the continuous input,
$S_n$ a sampling operator, $\mathcal{B}_n$ the discrete backbone,
and $\mathcal{B}$ its continuous counterpart. We seek
\begin{equation}
\left\|
\mathcal{B}_n(S_n u)-S_n\mathcal{B}(u)
\right\|_{\infty}
\longrightarrow 0
\end{equation}
as the discretization is refined. This expresses asymptotic
consistency and does not require exact equality at finite
resolutions. Although commonly termed discretization invariance in the literature, this property is more precisely understood as discretization consistency~\cite{calvello2025continuum,xiao2026functional,yang2026simple}. For consistency with literature, we use the term discretization-invariant.

\paragraph{Basis projection.}
Let $M_n=\operatorname{diag}(w_1,\ldots,w_n)$ contain
consistent quadrature weights; for uniform sampling of
a normalized domain, $w_i=1/n$. Define
\begin{equation}
\widetilde Q_n=\Phi_n^\top M_nQ_n,\quad
\widetilde K_n=\Psi_n^\top M_nK_n,\quad
\widetilde V_n=\Psi_n^\top M_nV_n.
\end{equation}
For shared continuous feature and basis functions,
these coefficients converge to discretization-independent
integrals, e.g.,
\begin{equation}
[\widetilde Q_n]_{j,:}
\longrightarrow
\int_{\Omega}\phi_j(x)q(x)^\top\,d\mu(x).
\end{equation}

\paragraph{TTT and reconstruction.}
With fixed basis size, initialization, learning rates,
and number of gradient steps, TTT is a continuous map
of these coefficients. Consequently,
\begin{equation}
W_n^{(\tau)}\longrightarrow W^{(\tau)},
\qquad
y_n(x)=\phi(x)^\top\widetilde Q_nW_n^{(\tau)}
\longrightarrow y(x).
\end{equation}
Finite compositions with continuous pointwise layers
preserve this convergence. Thus, the ReDT backbone is
discretization-consistent, since its tokenization and
positional representations are also consistent across
resolutions.
Importantly this argument does not establish invariance to adding new views or guarantee the same property for the retained prediction heads.

\subsection{Benchmarking Details}

\paragraph{Baselines.} In this study, we considered two general classes of 3D reconstruction models: base foundation models and training-free efficient models. The base foundation models include VGGT~\citep{wang2025vggt}, VGGT-Omega~\citep{wang2026vggtomega}, $\pi^3$~\citep{wang2026pi}, and DepthAnything3~\citep{depthanything3}. The efficient variants are all derived from the base VGGT backbone foundation models include FastVGGT~\citep{shen2026fastvggt}, FlashVGGT~\citep{wang2026flashvggt}, and ZipMap~\citep{jin2026zipmap}. Where applicable, we used the hyperparameter settings recommended by the original authors. All baselines (including efficient variants) were fine-tuned from their publicly available checkpoints using the same training protocol as ReDT, with early stopping based on validation loss, to ensure a fair and consistent comparison.

\paragraph{Benchmark Codebase.} Our benchmarking is adapted from MapAnything's publicly available benchmark code. The benchmark has been adapted for additional dataset processing and efficient ground truth postprocessing. Additionally, timing code was adapted from FastVGGT code base. 

\paragraph{Datasets.} 

We evaluate across a diverse set of real and synthetic multi-view benchmarks spanning indoor and outdoor scenes, static and dynamic content, and a wide range of native image resolutions. DynamicReplica provides synthetic dynamic indoor scenes with dense depth supervision at \(1280\times720\), while SAIL-VOS3D evaluates dynamic and cluttered indoor/outdoor sequences at \(1280\times800\). MVS-Synth and Spring provide synthetic FHD sequences at \(1920\times1080\), covering urban multi-view stereo and dynamic stereo settings, respectively. ScanNet++ evaluates real high-fidelity indoor reconstruction using registered imagery and laser/mesh geometry at \(1752\times1168\). To assess higher-resolution generalization, we additionally use UnrealStereo4K, which provides synthetic 4K stereo imagery at \(3840\times2160\), and ETH3D, which contains real high-resolution indoor and outdoor imagery with laser-scanner ground truth at resolutions exceeding \(4\)K.

\paragraph{Resolution Sweeps.}

We express image resolution as width $\times$ height in pixels.
For reference, the standard $16{:}9$ resolution formats are:

\begin{center}
\begin{tabular}{lll}
\hline
Label & Description & Standard Resolution \\
\hline
1K (HD) & High Definition & $1280 \times 720$ \\
2K (FHD) & Full High Definition & $1920 \times 1080$ \\
4K (QHD) & Quad High Definition & $2560 \times 1440$ \\
8K (UHD) & Ultra High Definition & $3840 \times 2160$ \\
\hline
\end{tabular}
\end{center}

These formats provide reference dimensions; images with other
aspect ratios may have different widths or heights.
We therefore specify the actual input dimensions when describing
evaluation settings. The base resolution denotes the starting
resolution of a sweep and is specified separately for each
experimental setting.

\subsection{Qualitative Outputs}

We compare ReDT with VGGT, VGGT-$\Omega$, and ZipMap on static HD sequences from ETH3D and dynamic HD sequences captured from Marvel’s Spider-Man 2 and Marvel’s Spider-Man Remastered~\citep{insomniac2023spiderman2}, evaluating depth maps, camera trajectories, and reconstructed point clouds. The Spider-Man sequences are captured at HD resolution and sampled at 10 FPS. On ETH3D, all methods recover similar overall scene layouts and camera trajectories. ReDT preserves distinct depth boundaries around the foreground furniture and background windows, while producing a coherent reconstruction of the courtyard and building façade. These results demonstrate that ReDT maintains reconstruction quality in static scenes.

The differences are more pronounced in the dynamic Spider-Man sequence. VGGT produces inconsistent depth within the moving character, an irregular camera trajectory, and a severely dispersed point cloud. ZipMap separates the character from the background more clearly, but loses the thin web strand and exhibits local irregularities in its trajectory. ReDT preserves both the character silhouette and the web strand, together with sharper background depth transitions. Its smooth camera trajectory and well-organized building geometry are visually comparable to VGGT-$\Omega$, with clearer recovery of fine foreground detail and the recessed background region in the displayed depth map. Together, these observations suggest that ReDT handles dynamic foreground content effectively while retaining coherent background geometry and camera motion estimates.


\begin{figure}[t]
\centering
\includegraphics[width=0.99\textwidth]{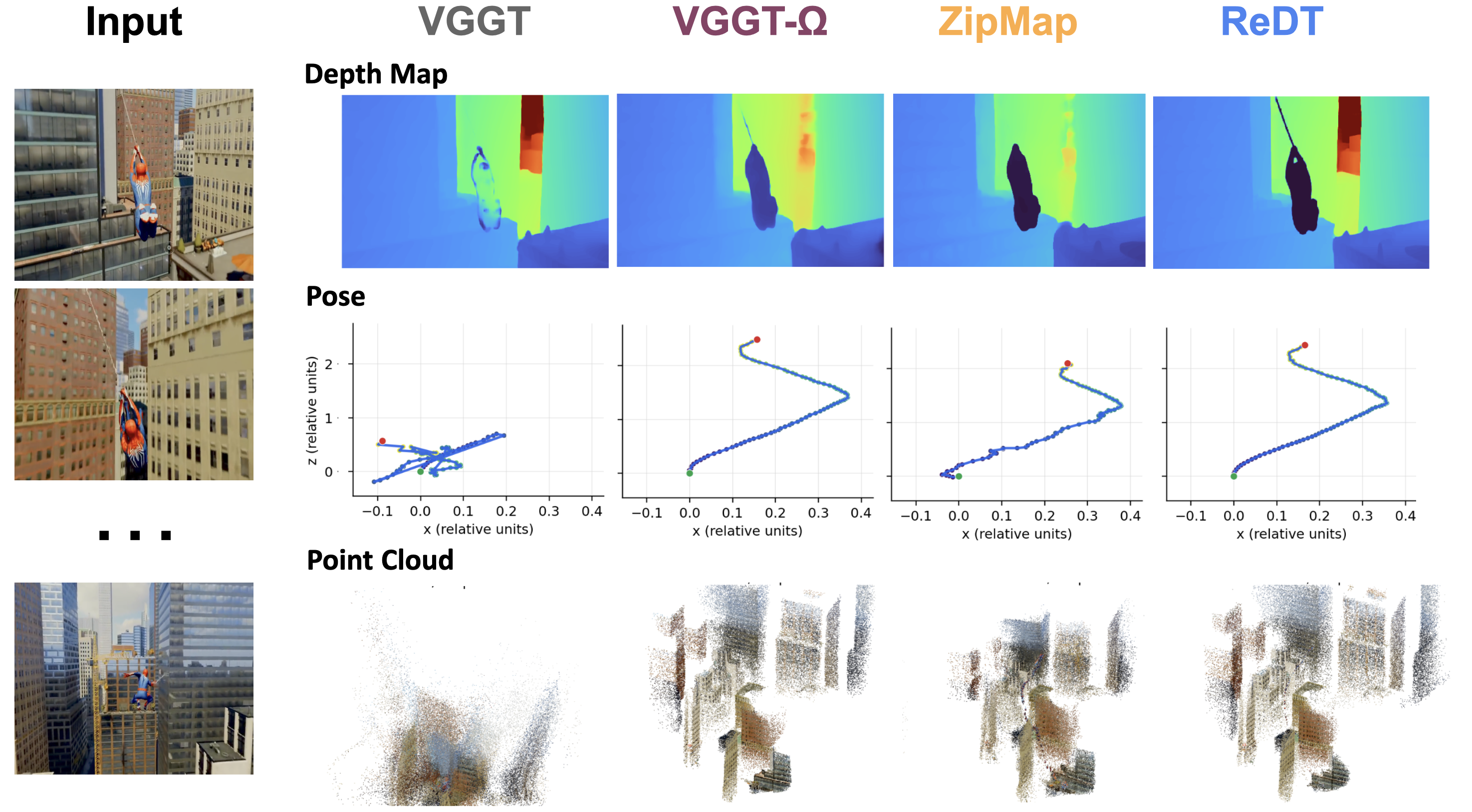}
\caption{Qualitative comparison on a dynamic Spider-Man sequence, showing depth maps, estimated camera trajectories, and reconstructed point clouds.}
\end{figure}

\begin{figure}[t]
\centering
\includegraphics[width=0.99\textwidth]{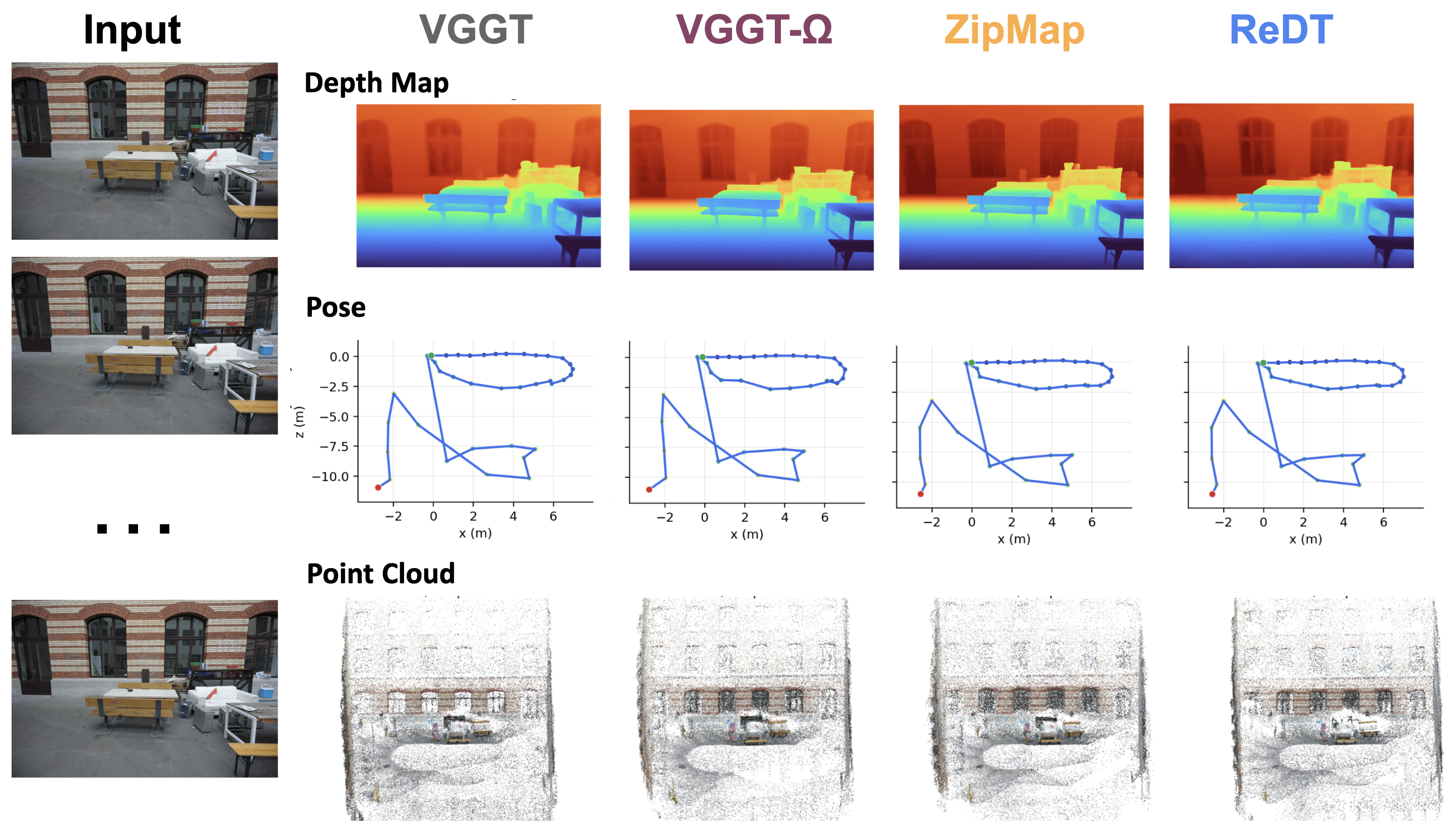}
\caption{Qualitative comparison on a static ETH3D scene, showing depth maps, estimated camera trajectories, and reconstructed point clouds.}
\end{figure}